\documentclass[11pt]{article}

\usepackage[margin=1in]{geometry}
\usepackage{amsmath,amssymb}
\usepackage{graphicx}
\usepackage{booktabs}
\usepackage{float}
\usepackage[hidelinks]{hyperref}

\title{TML-bench: Benchmark for Data Science Agents on Tabular ML Tasks}
\author{Mykola Pinchuk, PhD\\Independent Researcher\\San Jose, USA\\\texttt{pinchumkykola@gmail.com}}
\date{March 4, 2026}

\graphicspath{{figures/}}

\begin{document}
\maketitle

\begin{abstract}
Autonomous coding agents can produce strong tabular baselines quickly on Kaggle-style tasks. Practical value depends on end-to-end correctness and reliability under time limits. This paper introduces TML-bench, a tabular benchmark for data science agents on Kaggle-style tasks. This paper evaluates 10 OSS LLMs on four Kaggle competitions and three time budgets (240s, 600s, and 1200s). Each model is run five times per task and budget. A run is successful if it produces a valid submission and a private-holdout score on hidden labels that are not accessible to the agent. This paper reports median performance, success rates, and run-to-run variability. MiniMax-M2.1-TEE achieves the best aggregate performance score on all four competitions under the paper's primary aggregation. Average performance improves with larger time budgets. Scaling is noisy for some individual models at the current run count. Code and materials are available at \url{https://github.com/MykolaPinchuk/TML-bench/tree/master}.

\end{abstract}

\section{Introduction}
Tabular machine learning remains a common practical workload. The workflow includes data loading, feature preparation, model training, evaluation, iteration, and producing a correctly formatted submission artifact. Benchmarks that only test isolated coding tasks miss important failure modes and trade-offs.

This paper evaluates data science agents on a strict tabular benchmark with private-holdout scoring. The target is an auditable leaderboard built from repeatable runs. The benchmark considers score distributions across runs rather than a single best attempt.

For practitioners, this framing matters for two reasons. First, a good agent must be reliable. Capability on a single lucky run is insufficient. Second, comparisons should remain meaningful when tasks use different metrics and when runs are time-bounded.

\subsection*{Contributions}
This paper makes the following contributions:
\begin{itemize}
  \item This paper introduces a strict benchmark protocol for Kaggle-style tabular tasks with deterministic preparation, strict submission validation, and private-holdout scoring.
  \item This paper uses a repeatable reporting policy with fixed agent instructions, a fixed suite, and median-of-five aggregation with explicit coverage requirements.
  \item This paper adopts contamination controls and a low-cost evaluation setup intended to be runnable by individual practitioners (internet-off execution and pre-competition knowledge-cutoff model selection).
  \item This paper provides a reproducible evaluation protocol and supporting materials to regenerate figures and tables.
  \item This paper reports results and analysis that highlight performance, cross-competition consistency, reliability, and scaling with time budget.
\end{itemize}

\subsection*{Related work}
Several existing benchmarks evaluate coding-capable models and autonomous coding systems on programming and data workflows, but they differ in scope and evaluation philosophy.

SWE-bench focuses on real-world software engineering tasks derived from GitHub issues~[7]. It evaluates whether a system can produce code changes that resolve the issue under a test suite.

MLE-bench evaluates autonomous systems on ML engineering via Kaggle-style competitions curated from Kaggle~[8]. DSBench evaluates data science systems on a broader set of tasks that includes both data analysis and data modeling, and it introduces a metric designed to normalize heterogeneous task metrics~[9]. MLAgentBench evaluates autonomous coding systems on a suite of ML experimentation tasks in a controlled environment~[10].

Relative to MLE-bench, TML-bench makes a different set of trade-offs. MLE-bench optimizes breadth across many competitions and modalities. TML-bench instead constrains scope to tabular tasks and a fixed small suite, which reduces cross-modality confounding and makes run-to-run comparisons easier to interpret. Constraining to relatively small tabular datasets also keeps evaluation accessible: the marginal cost to run the full suite in the setup used for this paper was on the order of \$10. TML-bench also emphasizes repeatability with fixed instruction sets, explicit complete-coverage requirements, and repeated runs per setting. In addition, TML-bench uses private-holdout scoring outside the system workspace, keeps internet access disabled during runs, and selects evaluated models so their knowledge cutoff predates the start date of the tested Kaggle competitions. These design choices report reliability side by side with performance and reduce contamination risk from post-competition information.

\section{Benchmark and protocol}
\subsection{Suite and evaluation grid}
This paper evaluates model-assisted tabular ML work over a four-competition suite and three time budgets (240s, 600s, 1200s). The 1200s configuration uses an XGBoost-focused instruction set.

Each time budget uses a fixed instruction set, so the 240s/600s/1200s settings differ in both available time and the instructions given to the agent.

Reporting is restricted to models that reached complete five-run coverage across all 12 task$\times$budget settings (4 competitions $\times$ 3 time budgets). Under that rule, 10 models are included in the main tables.

\subsection{Prompt strategy and aggregation rule}
A fixed instruction template is used for all runs in this paper.

For each \texttt{(competition, model, budget)} setting, the reported value is the median of the earliest 5 successful runs ordered by \texttt{created\_at}.

\subsection{Evaluation logging and reproducibility}
The evidence in this paper is derived from logged run outcomes (status, score, runtime, and run configuration). I focus only on models with complete coverage: five successful runs for every task$\times$budget setting included in the benchmark.

\subsection{Agent harness (Kilo Code)}
Each run is executed in a clean, per-run workspace managed by the Kilo Code harness. The harness enforces the time budget, validates the submission format, and scores submissions on hidden holdout labels (not accessible to the agent).

Appendix~\ref{sec:appendix-harness} describes what Kilo Code is and why this paper standardizes on it.

\subsection{Contamination controls}
This paper uses two controls to reduce contamination risk from external information. First, internet access is disabled during benchmark runs. Second, evaluated models are selected so that their pretraining knowledge cutoff predates the start date of each tested Kaggle competition. In this suite, the earliest competition release is in late October 2025, and all listed cutoff dates in Appendix~\ref{sec:appendix-models} precede that date.

\subsection{Metrics and normalization}
Each competition has a task-defined metric. This paper reports \texttt{score\_raw} in the task's native direction (for example, AUC where higher is better, and RMSE where lower is better).

Raw metric values are not comparable across competitions because they have different scales and different directions. To build aggregate leaderboards and scaling plots, this paper uses a within-setting min-max normalization:
\begin{itemize}
  \item For each \texttt{(competition, budget)} setting, compute each model's five-run median \texttt{score\_raw}.
  \item Convert to a common ``higher is better'' value: \texttt{value\_hib = score\_raw} for higher-is-better metrics, and \texttt{value\_hib = -score\_raw} for lower-is-better metrics.
  \item Min-max normalize within that setting so the best model gets 1.0 and the worst gets 0.0:
    \[
      \mathrm{score} = \frac{\mathrm{value\_hib} - \min(\mathrm{value\_hib})}{\max(\mathrm{value\_hib}) - \min(\mathrm{value\_hib})}.
    \]
\end{itemize}

Appendix~\ref{sec:appendix-method} defines the normalization and the primary aggregation precisely.

\subsection{Time budgets}
This paper evaluates three wall-clock time budgets per competition: 240 seconds, 600 seconds, and 1200 seconds. A time budget represents the total time available to the agent to read the task, train, iterate, and produce a final submission file.

Implementation details (what ``240s'' means in practice):
\begin{itemize}
  \item The time budget applies to the agent interaction stage, enforced as a hard timeout by the Kilo Code harness.
  \item Runtime is measured from run start to the end of the agent stage (or timeout). Validation and private-holdout scoring occur after this stage and do not count against the time budget.
  \item If the harness observes that the agent exceeded the time budget (with a small grace window for timing granularity), the run is marked as a timeout even if a submission file exists.
\end{itemize}

Time budgets are included because ``fast baseline'' and ``iterative improvement'' are different capabilities. A strict budget makes these trade-offs visible under a repeatable protocol.

\section{Results}
This section reports aggregate performance, cross-competition consistency, reliability and stability, scaling with time budget, and per-competition highlights. The primary aggregate leaderboard uses a normalization that enables comparisons across competitions and time budgets.

\subsection{Key findings}
\begin{itemize}
  \item \texttt{MiniMax-M2.1-TEE} achieves the best aggregate performance score on all four competitions under the paper's primary aggregation.
  \item Reliability varies meaningfully even among strong performers. Success-rate and stability plots show clear separation between more and less reliable models.
  \item Some models improve substantially with larger time budgets, while other models remain relatively flat. Marginal-gain and monotonicity views capture these patterns.
\end{itemize}

\subsection{Aggregate performance leaderboard}
The aggregate leaderboard is derived from five-run medians and normalized via within-setting min-max scaling so that scores from different competitions (AUC vs RMSE) are comparable.

The method is as follows:
\begin{itemize}
  \item The unit of aggregation is the model's five-run median \texttt{score\_raw} for each \texttt{(competition, budget)} setting.
  \item Convert each setting to a common ``higher is better'' direction: \texttt{value\_hib = score\_raw} for higher-is-better metrics and \texttt{value\_hib = -score\_raw} for lower-is-better metrics.
  \item Within each setting, apply min-max normalization so that the best model gets 1.0 and the worst model gets 0.0:
    \[
      \mathrm{score} = \frac{\mathrm{value\_hib} - \min(\mathrm{value\_hib})}{\max(\mathrm{value\_hib}) - \min(\mathrm{value\_hib})}.
    \]
  \item The paper's performance score (the leaderboard value) uses the ``best budget per competition'' aggregation: for each \texttt{(model, competition)}, take the best normalized score across the three budgets, then average across the four competitions with equal weights.
\end{itemize}

This normalization is applied after accounting for metric direction (for example, AUC is higher-is-better, while RMSE is lower-is-better). Unlike rank-based scoring, min-max normalization preserves absolute metric gaps linearly within each \texttt{(competition, budget)} setting.

These scores are relative within each \texttt{(competition, budget)} setting, because they depend on the observed range of model performance in that setting.

\begin{figure}[t]
  \centering
  \includegraphics[width=\linewidth]{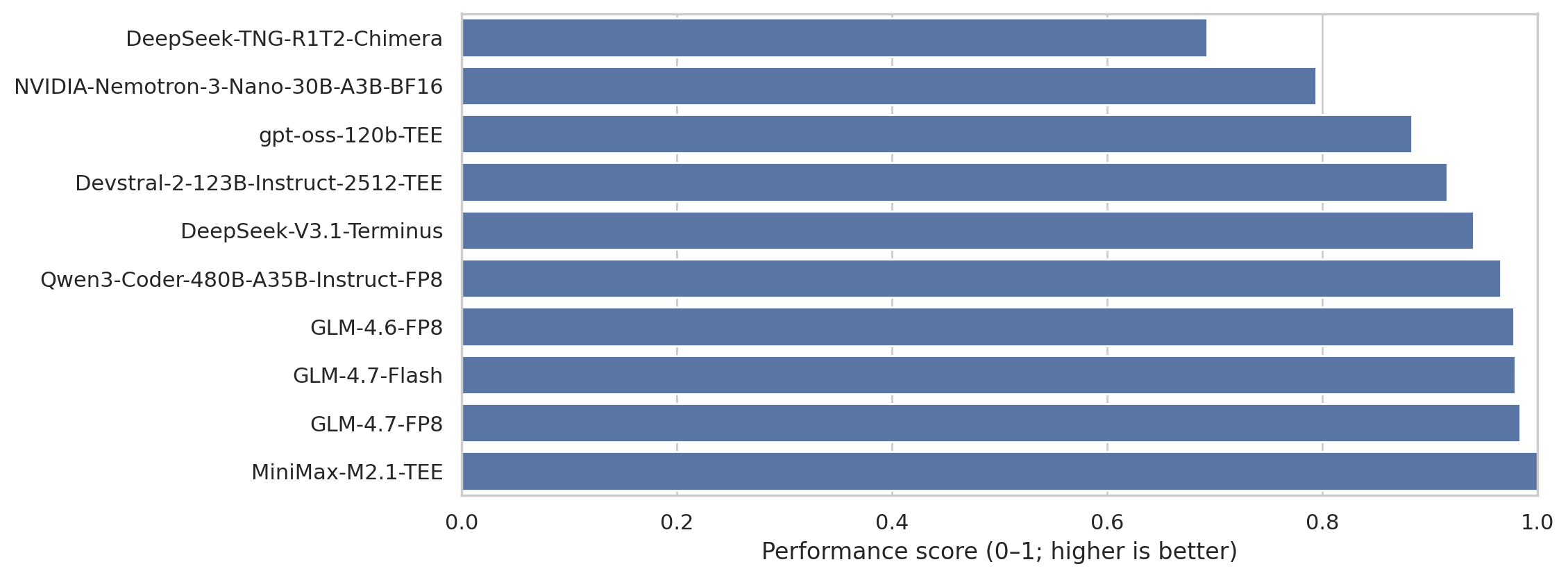}
  \caption{Aggregate performance leaderboard (primary aggregation: best budget per competition).}
  \label{fig:primary-leaderboard}
\end{figure}

Figure~\ref{fig:primary-leaderboard} is the main comparison figure in this paper. Robustness variants (secondary) include: (i) an overall aggregation that averages all \texttt{(competition, budget)} settings equally, and (ii) a 1200s-only aggregation. See Appendix~\ref{sec:appendix-robustness}.

\subsection{Cross-competition consistency}
\begin{figure}[t]
  \centering
  \includegraphics[width=\linewidth]{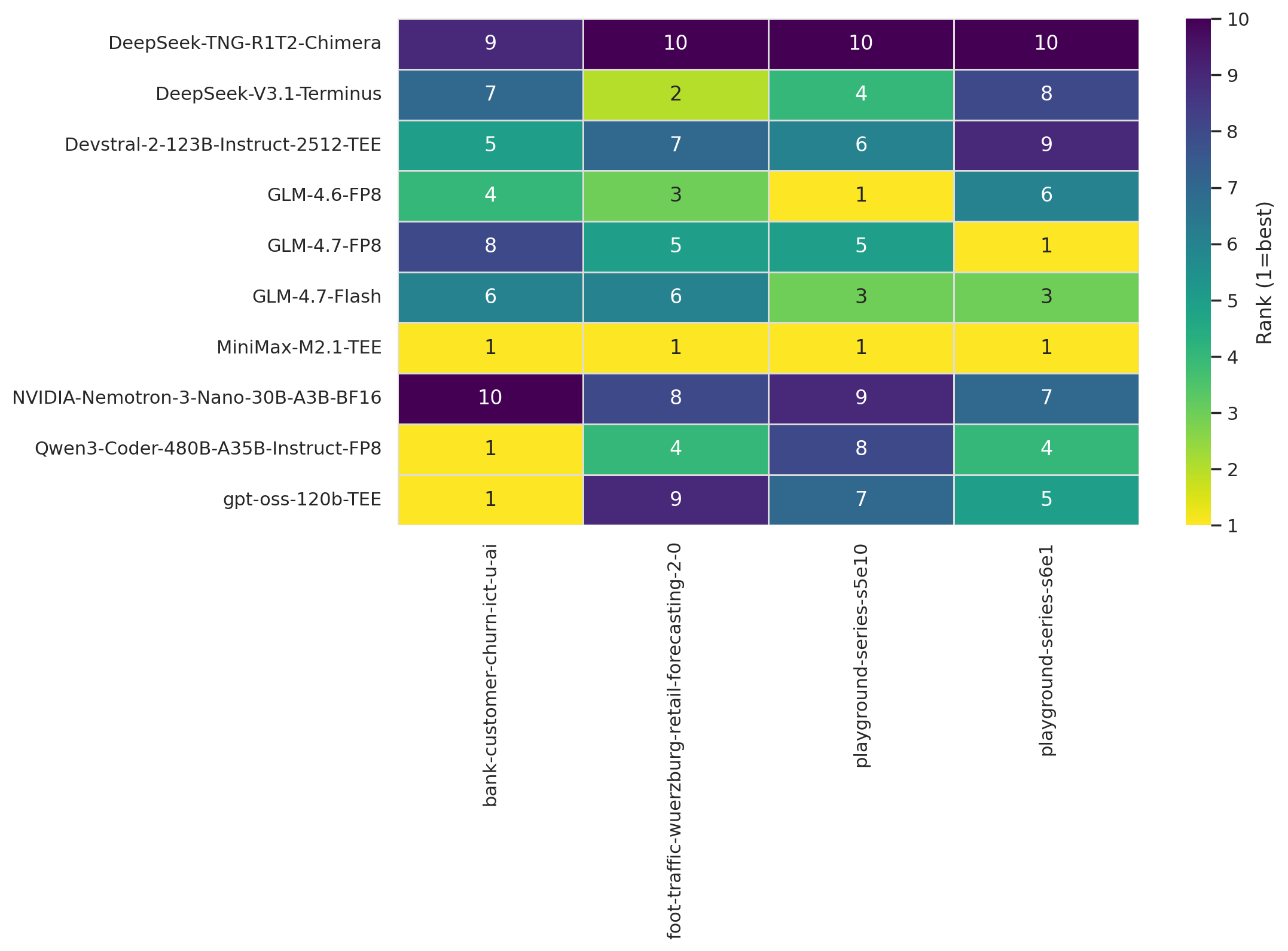}
  \caption{Per-competition ranks (1=best).}
  \label{fig:consistency-heatmap}
\end{figure}

Per-competition ranks are computed in the same normalized space as the primary aggregate leaderboard (best budget per competition). Figure~\ref{fig:consistency-heatmap} shows each model's rank (1=best) per competition.

Figure~\ref{fig:consistency-heatmap} keeps rank information as a compact view of cross-competition ordering. Rank variability across competitions is summarized via rank standard deviation (lower is more consistent). See Appendix~\ref{sec:appendix-consistency}.

\subsection{Reliability and stability}
Reliability has two components: run success rate (how often a run yields a valid score) and within-setting stability (how variable a model is across the five runs used for each reported setting). The trade-off is summarized via a Pareto-style plot (performance vs stability; dot fill color indicates success rate). Each dot corresponds to one model and is labeled with a callout pointer.

\begin{figure}[t]
  \centering
  \includegraphics[width=\linewidth]{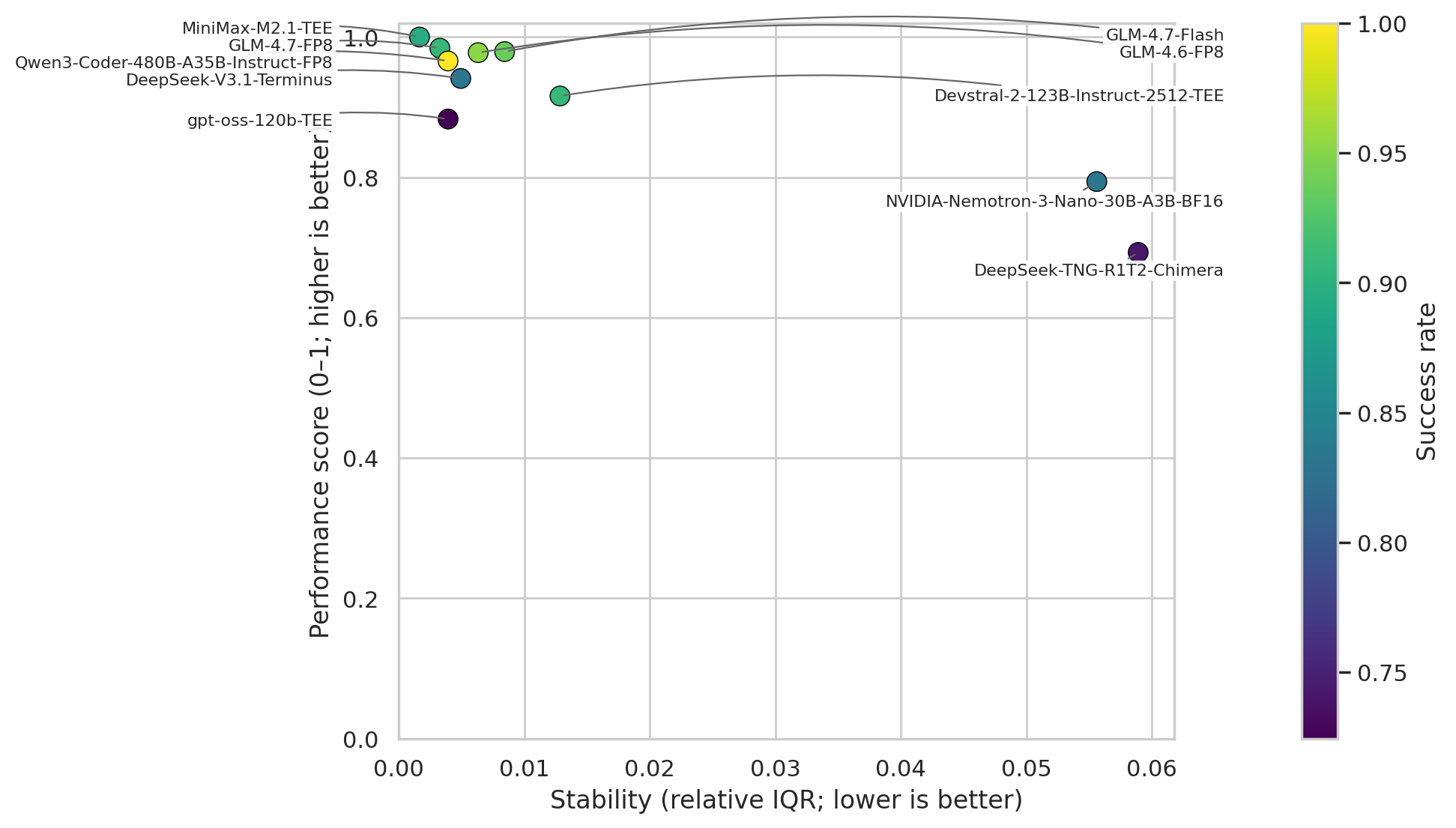}
  \caption{Performance vs stability (each dot is one model; fill color indicates success rate).}
  \label{fig:reliability-pareto}
\end{figure}

Figure~\ref{fig:reliability-pareto} highlights the reliability-performance trade-off. Supporting breakdown plots for success rate and stability are included in Appendix~\ref{sec:appendix-reliability}.

\subsection{Scaling with time budget}
Normalized performance is analyzed as time budget increases from 240s to 600s to 1200s, averaged across the four competitions.

On aggregate, scaling is broadly consistent with the expected monotonic pattern. This paper defines a model's per-competition scaling curve as monotone if its direction-corrected five-run medians do not worsen as budget increases (240s $\rightarrow$ 600s $\rightarrow$ 1200s). Under this definition, 23/40 (57.5\%) of model$\times$competition curves are monotone, and the median model is monotone in 62.5\% of competitions. Appendix~\ref{sec:appendix-scaling} reports monotonicity rates and marginal gains.

\begin{figure}[t]
  \centering
  \includegraphics[width=\linewidth]{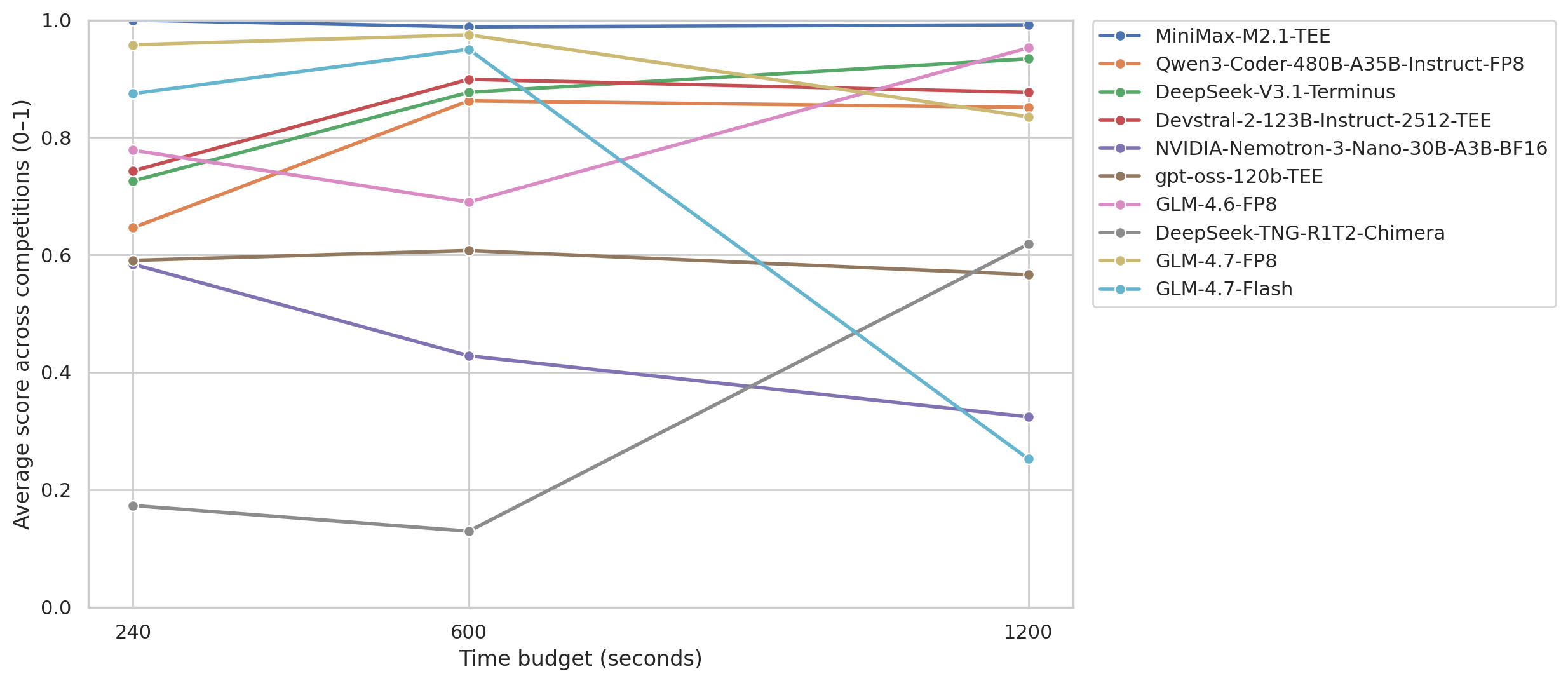}
  \caption{Scaling with time budget.}
  \label{fig:scaling-main}
\end{figure}

Figure~\ref{fig:scaling-main} summarizes budget scaling in normalized score space. At the individual model level, scaling can be noisy. Each task$\times$budget setting is summarized by five successful runs. More runs are likely required for stable model-level scaling curves. Appendix~\ref{sec:appendix-scaling} reports marginal gains and monotonicity rates.

\subsection{Per-competition highlights}
This subsection highlights a small number of representative task/budget settings. Full five-run tables are available in the companion repository materials.

\paragraph{bank-customer-churn-ict-u-ai (AUC; higher is better).}
The strongest median AUC at 1200s is 0.928000 (GPT OSS 120B TEE).
At 240s, the top median is 0.926671 (MiniMax-M2.1-TEE).
This task shows top-tier clustering in the 0.92x range, with notable underperformance from some models in specific settings (for example, NVIDIA-Nemotron-3-Nano at 0.813105 at 1200s).

\paragraph{foot-traffic-wuerzburg-retail-forecasting-2-0 (RMSE; lower is better).}
MiniMax-M2.1-TEE is best at all three budgets (0.066846, 0.065770, 0.065489).
A key instability signal appears for GLM 4.7 Flash at 1200s: median 0.107502 with IQR 0.070186..0.221725. This IQR is substantially wider than neighboring models.

\paragraph{playground-series-s5e10 (RMSE; lower is better).}
At 1200s, medians are tightly clustered near 0.0562, with the best setting at 0.056190 (GLM-4.6-FP8) and many models within a few $1\mathrm{e}{-4}$.

\paragraph{playground-series-s6e1 (RMSE; lower is better).}
MiniMax-M2.1-TEE leads at 1200s with RMSE 8.699779.
At 600s, TNG-R1T2-Chimera has a large failure-mode outlier (median 10.199380, IQR 9.088197..13.444163). This motivates caution when interpreting single-budget standings.

\section{Stability notes}
The stability companion should be read jointly with median performance. Several settings show narrow IQRs, while others exhibit broad or asymmetric spread.

Examples of high-variance settings include:
\begin{itemize}
  \item NVIDIA-Nemotron-3-Nano on s6e1 at 240s: 9.054929 (9.043837..10.604385).
  \item DeepSeek-V3.1-Terminus on foot-traffic at 600s: 0.068627 (0.066899..0.166052).
\end{itemize}

\section{Limitations}
This paper reports results for 10 models with complete coverage under the paper's evaluation grid.

\subsection{Token accounting}
Token consumption is currently unavailable in the run logs used for this paper (only \texttt{max\_tokens} configuration is present). Token efficiency is deferred to a later revision after token and cost instrumentation is added.

\subsection{Budget scaling and run count}
Two constraints matter when interpreting scaling with time budget. First, budgets are coupled to instruction sets (the 1200s setting uses an XGBoost-focused instruction set), so scaling reflects both more time and different instructions. Second, each reported setting is based on five successful runs, which is enough to show aggregate patterns but can be noisy at the individual model level.

\section{Reproducibility materials}
This paper is accompanied by a repository that contains run logs, scripts to regenerate figures and tables from those logs, and validation steps to confirm coverage.

\section*{References}
\begin{enumerate}
  \item Kilo Code (website). \url{https://www.kilocode.app/} (accessed 2026-02-14).
  \item Kilo Code documentation. \url{https://kilo.ai/docs} (accessed 2026-02-14).
  \item Kilo Code GitHub organization. \url{https://github.com/Kilo-Org} (accessed 2026-02-14).
  \item OpenRouter rankings (``Top Apps'', weekly tokens, opt-in tracking). \url{https://openrouter.ai/rankings} (accessed 2026-02-14).
  \item OpenRouter documentation: App attribution and rankings. \url{https://openrouter.ai/docs/app-attribution} (accessed 2026-02-14).
  \item Chen, T., and Guestrin, C. XGBoost: A Scalable Tree Boosting System. KDD 2016. \url{https://doi.org/10.1145/2939672.2939785}.
  \item Jimenez, C., Yang, J., Wettig, A., et al. SWE-bench: Can Language Models Resolve Real-World GitHub Issues? (2023). \url{https://arxiv.org/abs/2310.06770}.
  \item Chan, J., Jain, N., Karampatziakis, N., et al. MLE-bench. (2024). \url{https://arxiv.org/abs/2410.07095}.
  \item Mart\'{i}nez-Gonz\'{a}lez, B., Sehgal, A., Gupta, A., et al. DSBench. (2024). \url{https://arxiv.org/abs/2409.07703}.
  \item Huang, Z., Yang, S., Zhou, J., et al. MLAgentBench. (2023). \url{https://arxiv.org/abs/2310.03302}.
\end{enumerate}

\clearpage
\appendix
\clearpage
\section{Models evaluated in this paper}
\label{sec:appendix-models}

Table~\ref{tab:model-inventory} lists the models included in the 10-model set evaluated in this paper and summarizes metadata useful for interpretation. Release dates, parameter counts, knowledge-cutoff disclosures, and license fields are taken from public model cards and announcements. Knowledge cutoff refers to pretraining cutoff.

\begin{table}[H]
\centering
\scriptsize
\setlength{\tabcolsep}{3pt}
\begin{tabular}{p{3.7cm}p{1.2cm}p{1.3cm}p{1.55cm}p{1.3cm}p{1.9cm}p{1.45cm}}
\toprule
Model & Provider & Type & Parameters & Release date & Knowledge cutoff & License \\
\midrule
Qwen3-Coder-480B-A35B & chutes & open weights & 480B total, 35B active & 2025-07-23 & 2025-01-22 & Apache-2.0 \\
gpt-oss-120b & chutes & open weights & 120B & 2025-08-05 & 2024-06-01 & Apache-2.0 \\
GLM-4.7-FP8 & chutes & open weights & 358B & 2025-12-22 & 2025-04 & MIT \\
GLM-4.7-Flash & chutes & open weights & 30B & 2025-12-22 & 2025-04 & MIT \\
MiniMax-M2.1-TEE & chutes & open weights & 230B total, 10B active & 2025-12-23 & 2025-06-25 & unknown \\
GLM-4.6-FP8 & chutes & open weights & 357B & 2025-09-30 & 2025-04 & MIT \\
DeepSeek-V3.1-Terminus & chutes & open weights & 671B total, 37B active & 2025-09-23 & 2024-07 & unknown \\
Nemotron-3-Nano-30B & chutes & open weights & 30B total, 3B active & 2025-12-15 & 2025-06-25 & NVIDIA Nemotron OMLA \\
Devstral-2-123B & chutes & open weights & 125B & 2025-12-09 & 2023-10-01 (alt 2024-02) & Modified MIT \\
DeepSeek-TNG-R1T2-Chimera & chutes & open weights & 671B & 2025-07-02 & \textasciitilde{}2024-07 & unknown \\
\bottomrule
\end{tabular}
\caption{Model inventory for the 10 models evaluated in this paper.}
\label{tab:model-inventory}
\end{table}

\clearpage
\section{Additional figures and robustness checks}
\label{sec:appendix-figures}

\subsection{Alternative aggregate score variants}
\label{sec:appendix-robustness}
\begin{figure}[t]
  \centering
  \includegraphics[width=\linewidth]{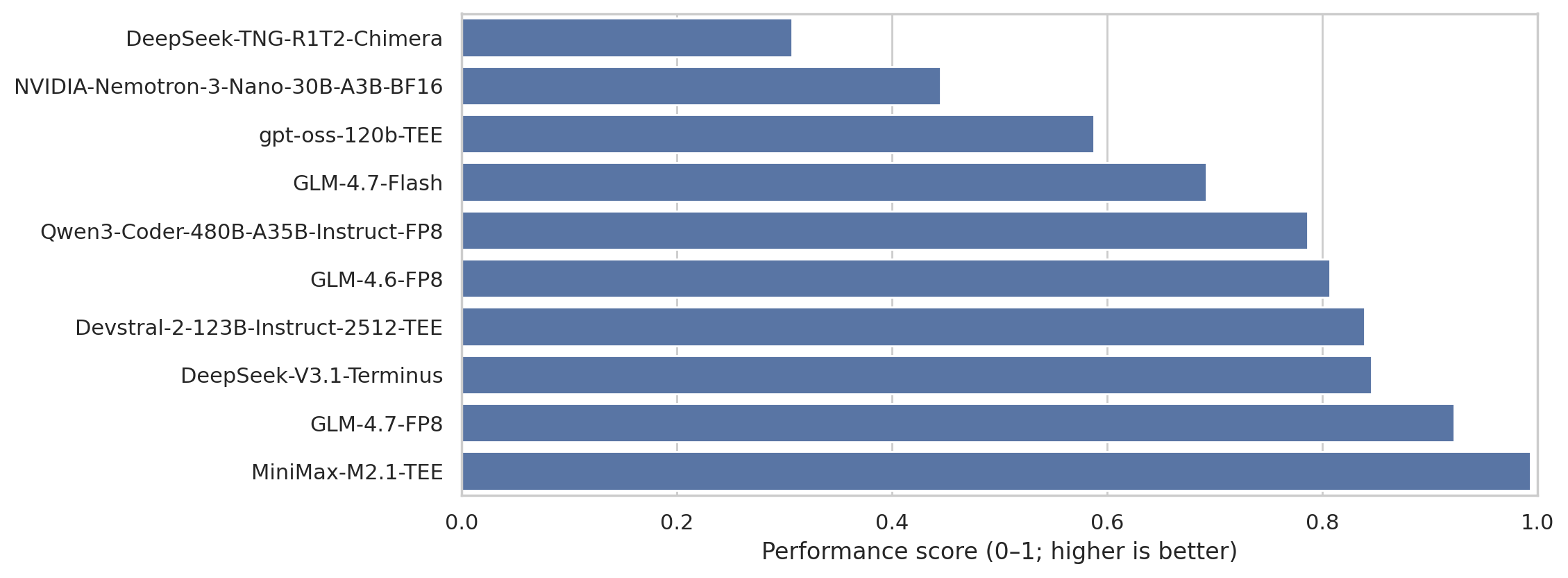}
  \caption{Overall aggregation (all competitions and all budgets).}
  \label{fig:appendix-overall-all-cells}
\end{figure}

\begin{figure}[t]
  \centering
  \includegraphics[width=\linewidth]{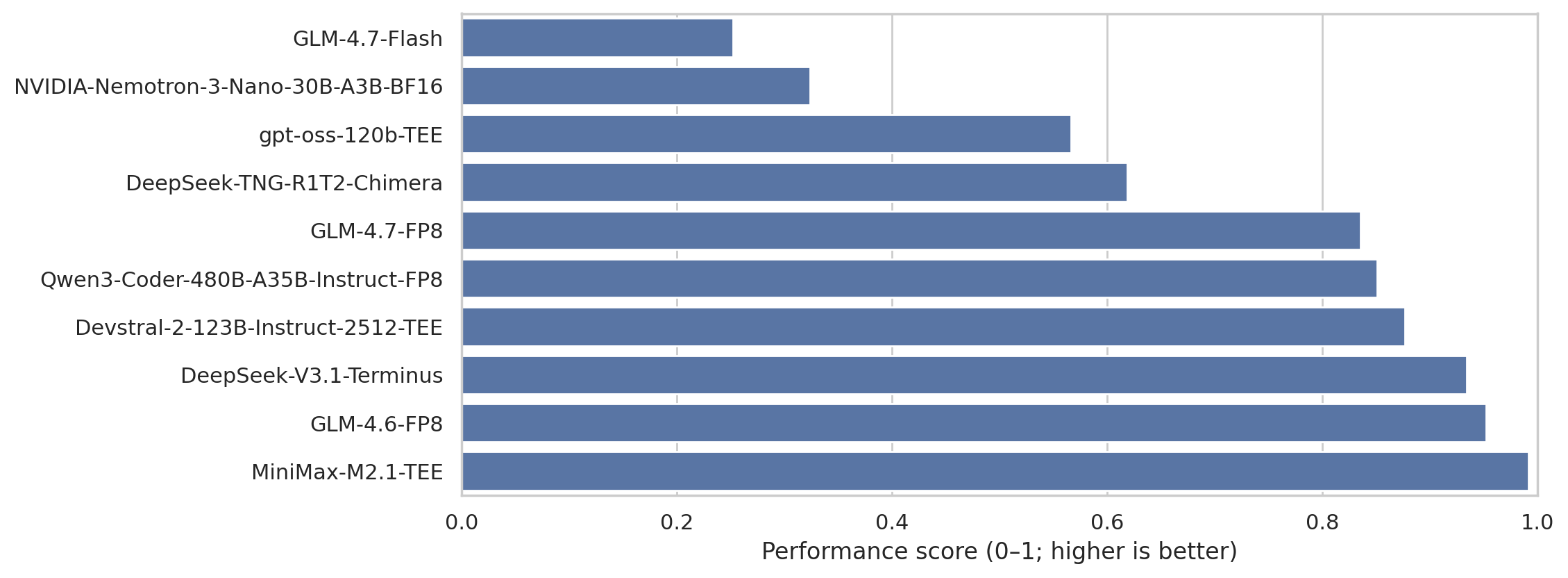}
  \caption{1200s-only aggregation.}
  \label{fig:appendix-1200-only}
\end{figure}

Figure~\ref{fig:appendix-overall-all-cells} and Figure~\ref{fig:appendix-1200-only} provide two alternative aggregate score views that complement the primary leaderboard in the main text.

\subsection{Supplementary rank-based views}
\label{sec:appendix-rank-based}
\begin{figure}[t]
  \centering
  \includegraphics[width=\linewidth]{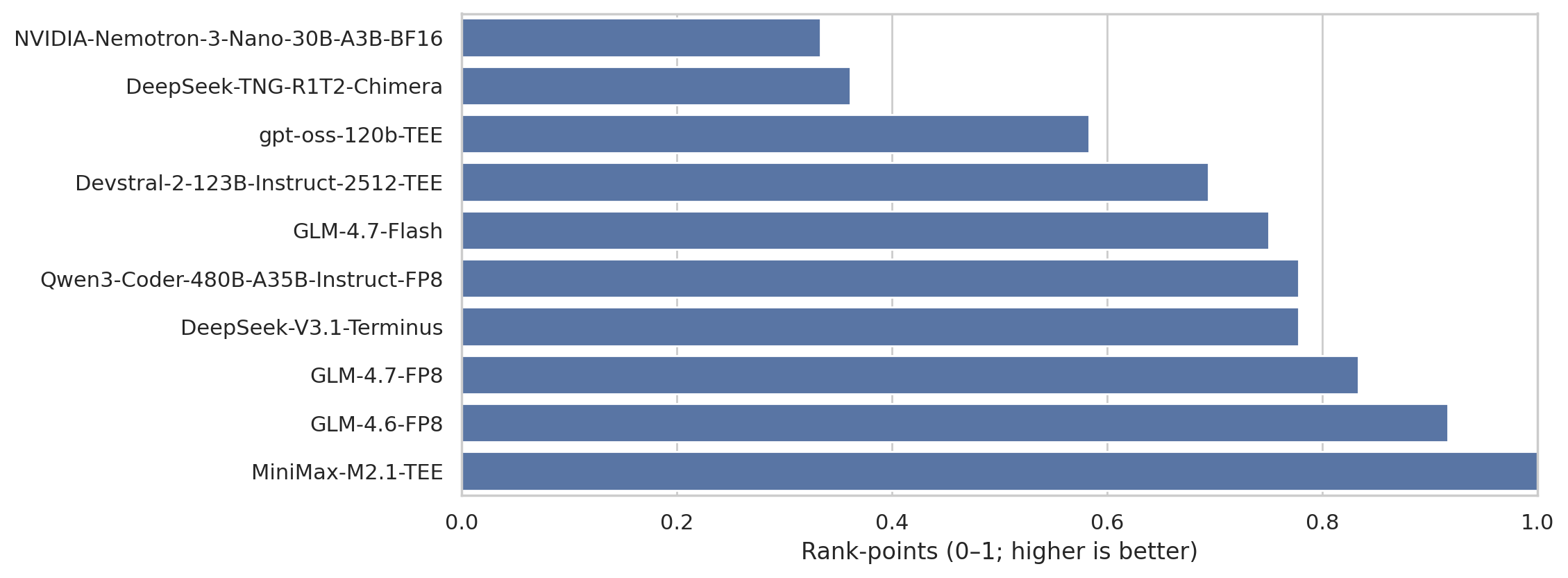}
  \caption{Rank-based aggregation (best budget per competition).}
  \label{fig:appendix-rank-best-budget}
\end{figure}

\begin{figure}[t]
  \centering
  \includegraphics[width=\linewidth]{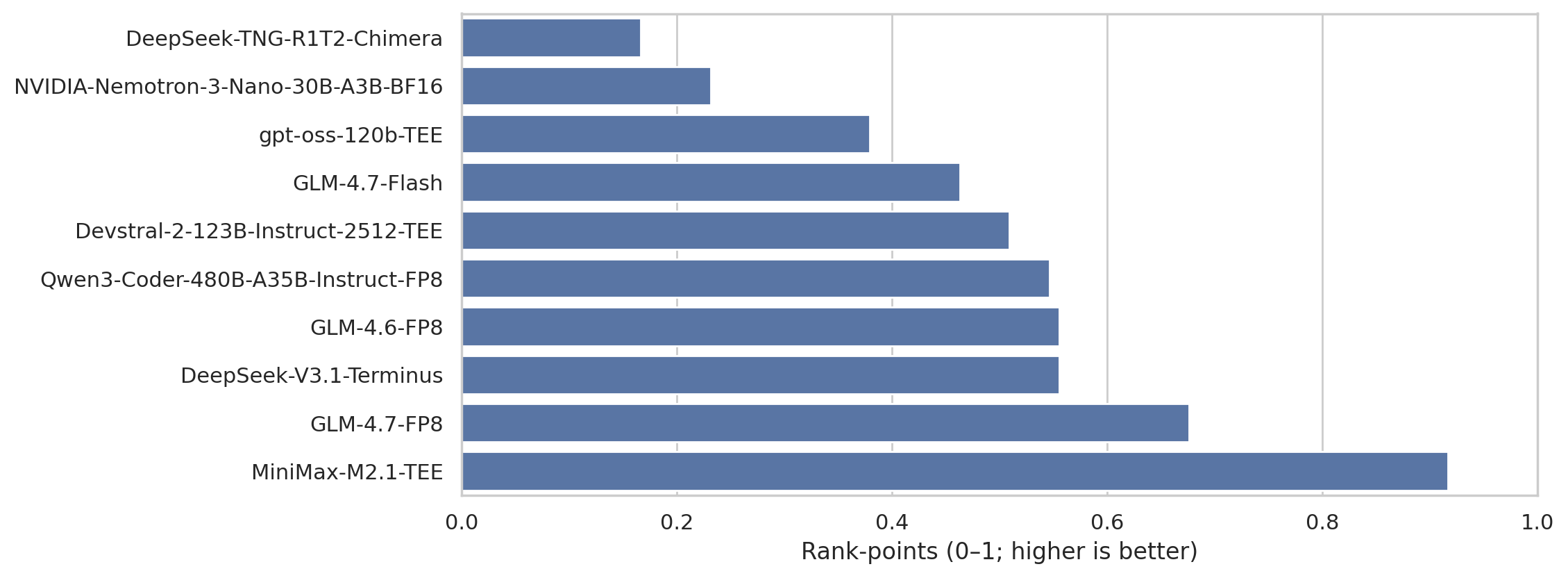}
  \caption{Rank-based aggregation (all competitions and all budgets).}
  \label{fig:appendix-rank-all-cells}
\end{figure}

\begin{figure}[t]
  \centering
  \includegraphics[width=\linewidth]{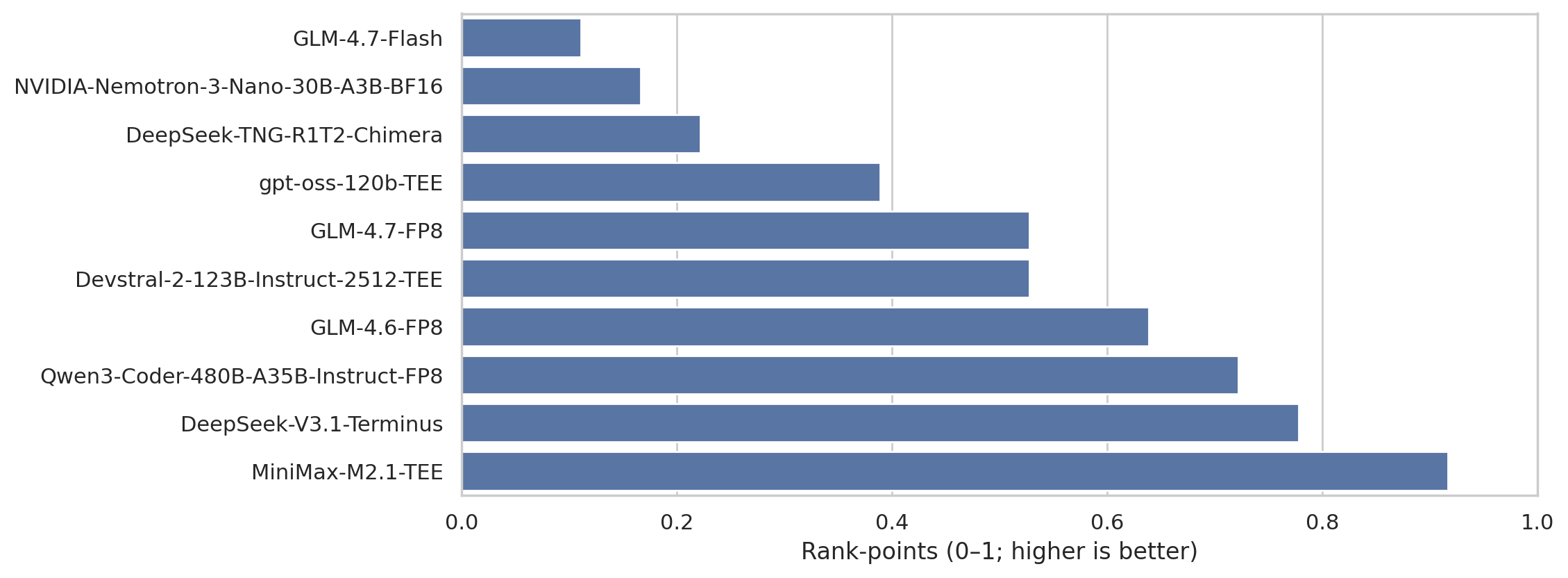}
  \caption{Rank-based aggregation (1200s only).}
  \label{fig:appendix-rank-1200-only}
\end{figure}

Figure~\ref{fig:appendix-rank-best-budget}, Figure~\ref{fig:appendix-rank-all-cells}, and Figure~\ref{fig:appendix-rank-1200-only} show the older rank-based normalization for comparison with the score-based normalization used in the main analysis.

\subsection{Consistency supplement}
\label{sec:appendix-consistency}
\begin{figure}[t]
  \centering
  \includegraphics[width=\linewidth]{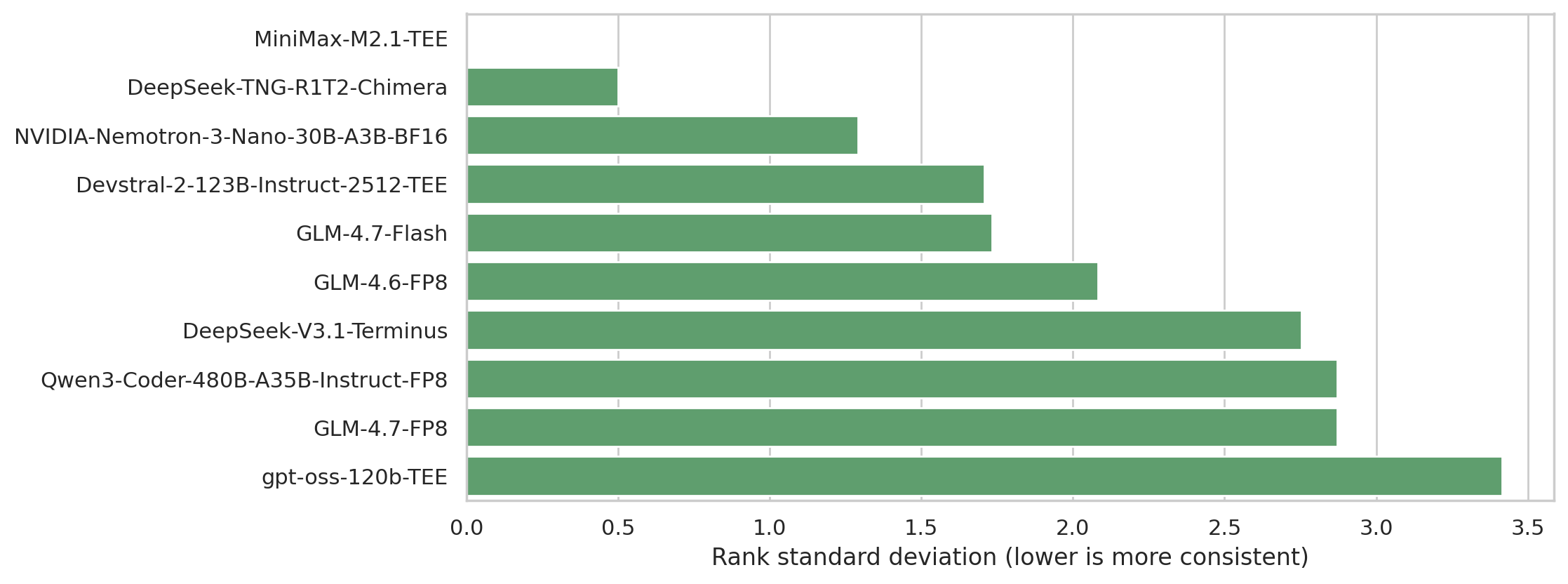}
  \caption{Rank standard deviation across competitions.}
  \label{fig:appendix-consistency-stddev}
\end{figure}

Figure~\ref{fig:appendix-consistency-stddev} reports rank variability across competitions (lower indicates higher consistency).

\subsection{Reliability and stability supplement}
\label{sec:appendix-reliability}
\begin{figure}[t]
  \centering
  \includegraphics[width=\linewidth]{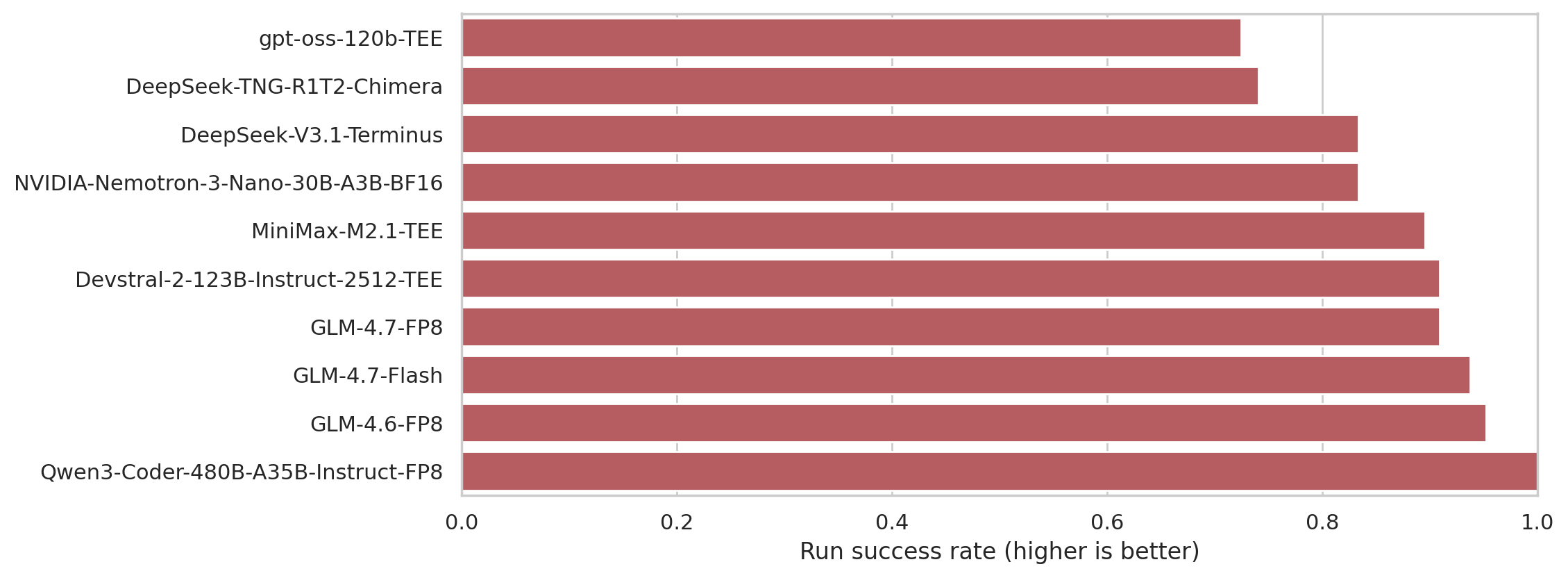}
  \caption{Run success rate.}
  \label{fig:appendix-success-rate}
\end{figure}

\begin{figure}[t]
  \centering
  \includegraphics[width=\linewidth]{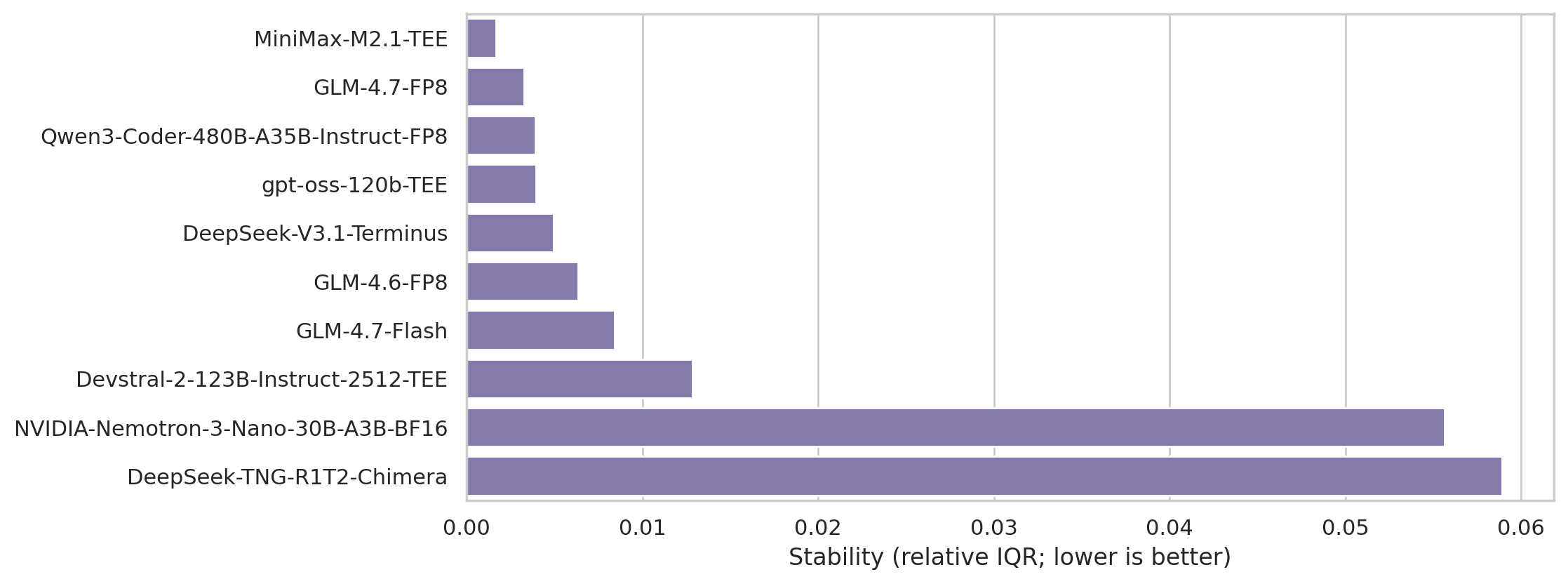}
  \caption{Stability via relative IQR.}
  \label{fig:appendix-stability-iqr}
\end{figure}

Figure~\ref{fig:appendix-success-rate} and Figure~\ref{fig:appendix-stability-iqr} separate reliability into completion rate and within-setting variability views.

\subsection{Scaling supplement}
\label{sec:appendix-scaling}
\begin{figure}[t]
  \centering
  \includegraphics[width=\linewidth]{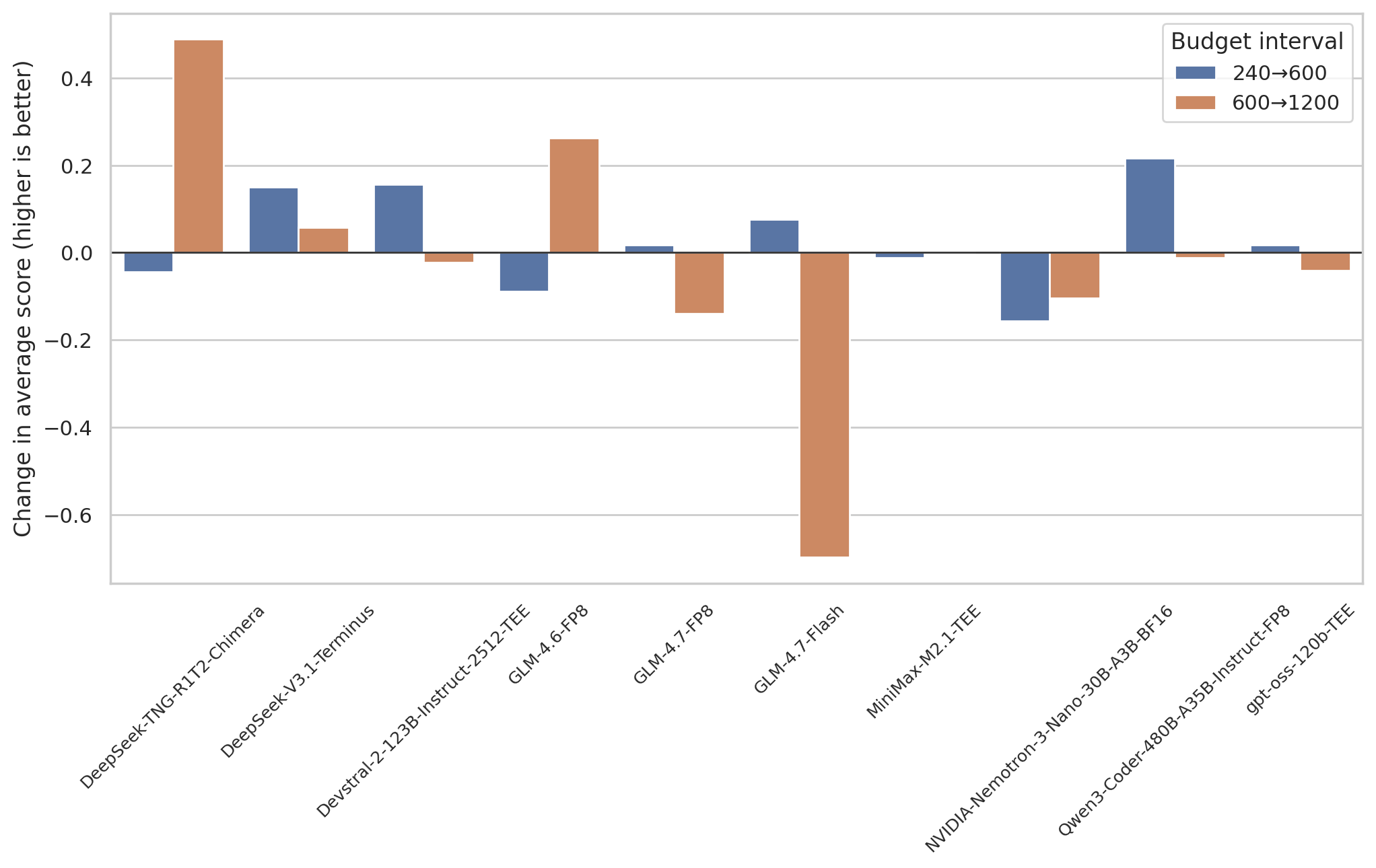}
  \caption{Marginal gains with increasing budget.}
  \label{fig:appendix-scaling-marginal}
\end{figure}

\begin{figure}[t]
  \centering
  \includegraphics[width=\linewidth]{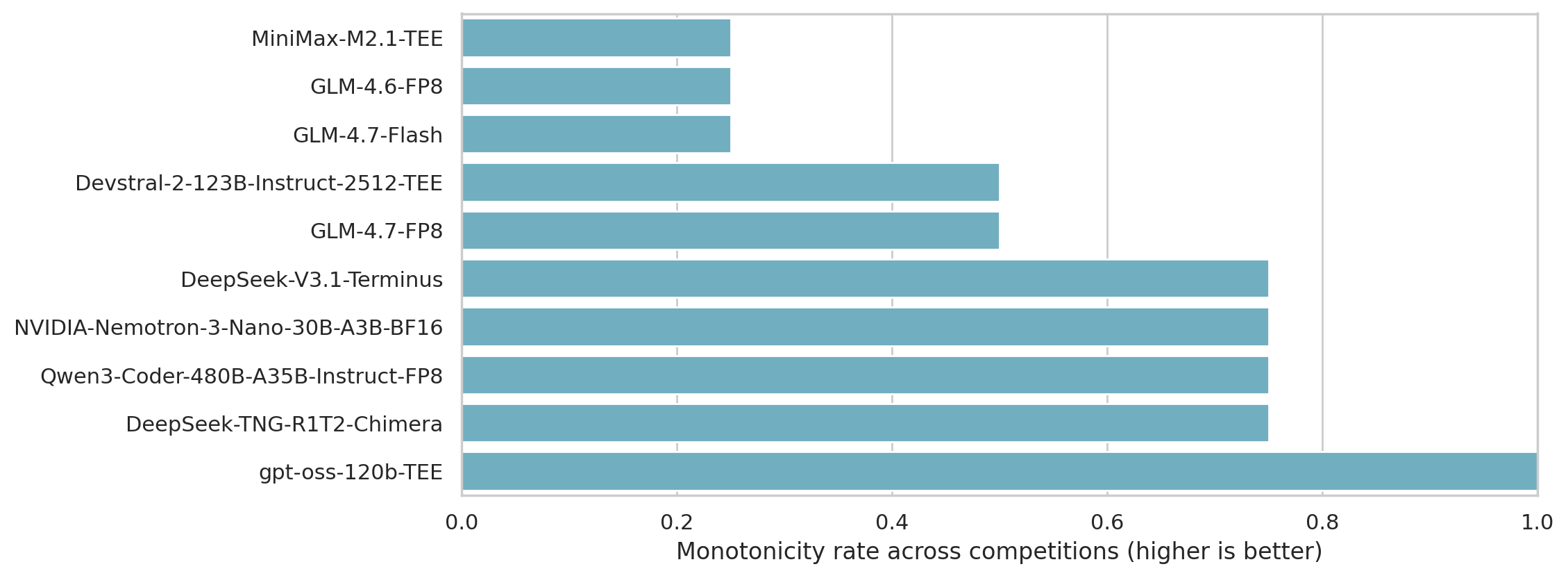}
  \caption{Monotonicity rate across budgets.}
  \label{fig:appendix-scaling-monotonicity}
\end{figure}

Figure~\ref{fig:appendix-scaling-marginal} and Figure~\ref{fig:appendix-scaling-monotonicity} provide supplementary scaling diagnostics.

\clearpage
\section{Scoring, aggregation, and normalization details}
\label{sec:appendix-method}

This appendix defines how scores are computed and how the aggregate leaderboard is constructed.

\subsection{Per-run scoring}
\begin{itemize}
  \item Each run produces a submission file.
  \item The harness validates the submission schema against the competition's expected format.
  \item The submission is scored on a private holdout set outside the agent workspace to produce \texttt{score\_raw} using the competition's metric (for example, AUC or RMSE).
\end{itemize}

\subsection{Per-setting aggregation}
This paper aggregates results for each \texttt{(competition, model, budget)} setting as follows:
\begin{itemize}
  \item Consider the earliest five successful runs.
  \item Report the median of their \texttt{score\_raw}.
\end{itemize}

\subsection{Min-max normalization}
Raw metrics are not directly comparable across competitions because they have different scales and directions. To build a single aggregate leaderboard, this paper uses within-setting min-max normalization.

This paper defines the normalized score within each \texttt{(competition, budget)} setting as follows:
\begin{enumerate}
  \item Convert median metrics to a common ``higher is better'' direction:
    \[
      \mathrm{value\_hib} =
      \begin{cases}
        \mathrm{score\_raw} & \text{if higher is better} \\
        -\mathrm{score\_raw} & \text{if lower is better.}
      \end{cases}
    \]
  \item Min-max normalize within the setting so that the best model receives 1.0 and the worst receives 0.0:
    \[
      \mathrm{score} = \frac{\mathrm{value\_hib} - \min(\mathrm{value\_hib})}{\max(\mathrm{value\_hib}) - \min(\mathrm{value\_hib})}.
    \]
\end{enumerate}

\subsection{Primary aggregation}
The primary aggregation is ``best budget per competition'':
\begin{itemize}
  \item For each \texttt{(model, competition)}, take the maximum normalized score across the three budgets.
  \item Average across the four competitions with equal weights.
\end{itemize}

\clearpage
\section{Harness details (Kilo Code)}
\label{sec:appendix-harness}

This appendix explains what Kilo Code is and why it is used as the single system harness in this paper.

\subsection{What Kilo Code is}
Kilo Code is an AI coding agent for VS Code. In TML-bench, it is used as the uniform interface between a model and the benchmark task workspace: the system reads task files, writes code, trains models, and produces a submission file.

Kilo Code is also widely used in practice. For example, OpenRouter's public ``Top Apps'' leaderboard (weekly tokens, based on opt-in app attribution) lists Kilo Code as \#2 as of 2026-02-14 (and as the highest-usage VS Code coding agent app in that list), see \url{https://openrouter.ai/rankings} and \url{https://openrouter.ai/docs/app-attribution}. This is a point-in-time snapshot; rankings vary over time.

\subsection{Why standardize on a single harness}
Many open-source harnesses exist, and different harnesses can introduce confounds: differences in tool availability, file access conventions, patch/apply mechanics, retry behavior, and failure handling. This paper standardizes on a single harness to reduce harness effects and make comparisons across models more interpretable.

Kilo Code was chosen for three practical reasons. First, it is reliable and mature enough to run repeatedly under timeouts and produce stable artifacts. Second, it is used with a wide range of open and API-served models, which reduces the risk of harness-model incompatibilities. Third, it enables fast sanity checks because it is available as a VS Code extension and basic end-to-end behavior can be verified quickly outside benchmark runs.

\subsection{What the harness enforces (high level)}
At a high level, the harness ensures that: (i) the system works in a clean per-run workspace with only the system-visible task inputs, (ii) the interaction stage is time-bounded (240s/600s/1200s), (iii) submissions are validated and normalized before scoring, and (iv) the final score is computed on hidden holdout labels outside the system workspace.

\subsection{Limitations of this choice}
Standardizing on Kilo Code improves comparability, but it narrows the scope of conclusions: results are about models as used through this harness under this protocol. Other harnesses may yield different absolute performance or failure rates.

\clearpage
\section{Operational lessons}
\label{sec:appendix-operational}

Running TML-bench reliably surfaced several operational lessons that may be useful to teams building similar agent evaluation systems.

\subsection{A control plane beats ad-hoc scripts}
Long-running suites benefit from a simple ``control plane'': durable run IDs, structured logs, and machine-readable events. This makes suites resumable, debuggable, and auditable. In contrast, ad-hoc shell scripts tend to fail silently, make partial failures hard to diagnose, and are difficult to parallelize safely.

\subsection{Circuit breakers save time and money}
When a provider or model enters a failure streak (timeouts, repeated invalid submissions, intermittent API errors), a circuit breaker can pause or skip that lane. This prevents burning compute and wall-clock on runs that are likely to fail again and allows the suite to make progress elsewhere.

\subsection{Incremental persistence prevents ``lost suites''}
Suites that take hours or days should write results continuously. Persisting every run outcome (including failures) as soon as it completes makes the overall process robust to interruptions: machine restarts, parent process crashes, or transient provider outages. It also supports incremental analysis rather than ``all-or-nothing'' end-of-suite reporting.

\subsection{Per-task resource caps improve stability}
Some tasks are more resource-sensitive than others (for example, heavy preprocessing, large feature matrices, or memory pressure under parallelism). Per-task caps (concurrency limits, memory limits, or stricter runtime limits) can prevent cascading failures and make the benchmark safer to run repeatedly.

\subsection{Post-run diagnostics are part of reliability}
A benchmark should treat ``why did this run fail?'' as a first-class output. Capturing a compact post-run diagnostic artifact (status, timeout vs.\ validation vs.\ runtime error, and a short trace or log pointer) turns failures into actionable debugging items and improves reproducibility of observed failure modes.

\clearpage
\section{Competition details}
\label{sec:appendix-competition-details}

This appendix summarizes the four benchmark competitions and their canonical public-input sizes used in this paper. Table~\ref{tab:competition-details} reports problem type, target, metric, and dataset dimensions from the generated \texttt{public/train\_public.csv} and \texttt{public/test\_public.csv} files.

\begin{table}[H]
\centering
\scriptsize
\setlength{\tabcolsep}{4pt}
\begin{tabular}{p{4.15cm}p{3.05cm}p{0.95cm}p{1.1cm}p{1.1cm}p{1.35cm}}
\toprule
Competition & Problem and target & Metric & Train rows & Test rows & Features \\
\midrule
bank-customer-churn-ict-u-ai & Binary classification: predict customer churn (\texttt{Exited}) & AUC & 12,000 & 3,000 & 12 \\
foot-traffic-wuerzburg-retail-forecasting-2-0 & Regression: forecast foot traffic (\texttt{target}) & RMSE & 41,155 & 10,289 & 4 \\
playground-series-s5e10 & Regression: predict road accident risk (\texttt{accident\_risk}) & RMSE & 414,203 & 103,551 & 12 \\
playground-series-s6e1 & Regression: predict exam scores (\texttt{exam\_score}) & RMSE & 504,000 & 126,000 & 11 \\
\bottomrule
\end{tabular}
\caption{Competition summary for the four tasks used in TML-bench. Feature counts exclude the \texttt{id} column and target column.}
\label{tab:competition-details}
\end{table}

\end{document}